\documentclass{article}

\usepackage[preprint]{nips_2018}

\usepackage[utf8]{inputenc} 
\usepackage[T1]{fontenc}    
\usepackage{hyperref}       
\usepackage{url}            
\usepackage{booktabs}       
\usepackage{amsfonts}       
\usepackage{nicefrac}       
\usepackage{microtype}      
\usepackage{graphicx}
\usepackage{bm}

\usepackage{xcolor}
\definecolor{darkgreen}{rgb}{0,0.6,0}
\definecolor{darkred}{rgb}{0.7,0.0,0}
\definecolor{darkblue}{rgb}{0,0.0,0.6}
\definecolor{darkpurple}{rgb}{0.4,0.0,0.6}

\title{PCA of high dimensional random walks with comparison to neural
network training}

\author{
  Joseph M.~Antognini\thanks{Work done as a Google AI Resident},
  Jascha Sohl-Dickstein \\
  Google Brain \\
  \texttt{\{antognini,jaschasd\}@google.com} \\
}

\def\kth{$k^{\textrm{{\scriptsize th}}}$}
\def\SOU{\textbf{S}_{\textrm{{\scriptsize OU}}}}
\def\XPCA{\textbf{X}_{\textrm{{\scriptsize PCA}}, k}}

\begin{document}

\maketitle

\begin{abstract}

  One technique to visualize the training of neural networks is to perform
  PCA on the parameters over the course of training and to project to the
  subspace spanned by the first few PCA components.  In this paper we
  compare this technique to the PCA of a high dimensional random walk.  We
  compute the eigenvalues and eigenvectors of the covariance of the
  trajectory and prove that in the long trajectory and high dimensional
  limit most of the variance is in the first few PCA components, and that
  the projection of the trajectory onto any subspace spanned by PCA
  components is a Lissajous curve.  We generalize these results to a random
  walk with momentum and to an Ornstein-Uhlenbeck processes (i.e., a random
  walk in a quadratic potential) and show that in high dimensions the walk
  is not mean reverting, but will instead be trapped at a fixed distance
  from the minimum.  We finally compare the distribution of PCA variances
  and the PCA projected training trajectories of a linear model trained on
  CIFAR-10 and ResNet-50-v2 trained on Imagenet and find that the
  distribution of PCA variances resembles a random walk with drift.

\end{abstract}

\section{Introduction}

Deep neural networks (NNs) are extremely high dimensional objects.  A
popular deep NN for image recognition tasks like ResNet-50 \citep{he+16} has
$\sim$25 million parameters, and it is common for language models to have
more than one billion parameters \citep{jozefowicz+16}.  This
overparameterization may be responsible for NNs impressive generalization
performance \citep{novak2018sensitivity}.  Simultaneously, the high
dimensional nature of NNs makes them very difficult to reason about.

Over the decades of NN research, the common lore about the geometry of the
loss landscape of NNs has changed dramatically.  In the early days of NN
research it was believed that NNs were difficult to train because they
tended to get stuck in suboptimal local minima.  \citet{dauphin+14} and
\citet{choromanska+15} argued that this is unlikely to be a problem for most
loss landscapes because local minima will tend not to be much worse than
global minima.  There are, however, many other plausible properties of the
geometry of NN loss landscapes that could pose obstacles to NN optimization.
These include: saddle points, vast plateaus where the gradient is very
small, cliffs where the loss suddenly increases or decreases, winding
canyons, and local maxima that must be navigated around.

Ideally we would like to be able to somehow visualize the loss landscapes of
NNs, but this is a difficult, perhaps even futile, task because it involves
embedding this extremely high dimensional space into very few dimensions ---
typically one or two.  \citet{goodfellow+15} introduced a visualization
technique that consists of plotting the loss along a straight line from the
initial point to the final point of training (the ``royal road'').  The
authors found that the loss often decreased monotonically along this path.
They further considered the loss in the space from the residuals between the
NN's trajectory to this royal road.  Note that while this is a
two-dimensional manifold, it is not a linear subspace.  \citet{lorch16}
proposed another visualization technique in which principal component
analysis (PCA) is performed on the NN trajectory and the trajectory is
projected into the subspace spanned by the lowest PCA components.  This
technique was further explored by \citet{li+18}, who noted that most of the
variance is in the first two PCA components.

In this paper we consider the theory behind this visualization technique.
We show that PCA projections of random walks in flat space qualitatively
have many of the same properties as projections of NN training trajectories.
We then generalize these results to a random walk with momentum and a random
walk in a quadratic potential, also known as an Ornstein-Uhlenbeck process
\citep{uhlenbeck+ornstein30}.  This process is more similar to NN
optimization since it consists of a deterministic component (the true
gradient) plus a stochastic component.  In fact, recent work has suggested
that stochastic gradient descent (SGD) approximates a random walk in a
quadratic potential \citep{ahn+12, mandt+16, smith+le17}.  Finally, we
perform experiments on linear models and large NNs to show how closely they
match this simplified model.

The approach we take to study the properties of the PCA of high dimensional
random walks in flat space follows that of \citet{moore+ahmed18}, but we
correct several errors in their argument, notably in the values of the
matrix $\textbf{S}^T \textbf{S}$ and the trace of $(\textbf{S}^T
\textbf{S})^{-1}$ in Eq.~\ref{eq:trace}.  We also fill in some critical
omissions, particularly the connection between banded Toeplitz matrices and
circulant matrices.  We extend their contribution by proving that the
trajectories of high dimensional random walk in PCA subspaces are Lissajous
curves and generalizing to random walks with momentum and Ornstein-Uhlenbeck
processes.

\section{PCA of random walks in flat space}
\label{sec:flat_space}

\subsection{Preliminaries}

Let us consider a random walk in $d$-dimensional space consisting of $n$
steps where every step is equal to the previous step plus a sample from an
arbitrary probability distribution, $\mathcal{P}$, with zero mean and a
finite covariance matrix.\footnote{The case of a constant non-zero mean
corresponds to a random walk with a constant drift term.  This is not an
especially interesting extension from the perspective of PCA because in the
limit of a large number of steps the first PCA component will simply pick
out the direction of the drift (i.e., the mean), and the remaining PCA
components will behave as a random walk without a drift term.}  For
simplicity we shall assume that the covariance matrix has been normalized so
that its trace is 1.  This process can be written in the form
\begin{equation}
  \textbf{x}_t = \textbf{x}_{t-1} + \bm{\xi}_t,
  \qquad \bm{\xi}_t \sim \mathcal{P},
\end{equation}
where $\textbf{x}_t$ is a $d$-dimensional vector and $\textbf{x}_0 =
\textbf{0}$.  If we collect the $\textbf{x}_t$s together in an $n \times d$
dimensional design matrix $\textbf{X}$, we can then write this entire
process in matrix form as
\begin{equation}
\label{eq:rw_matrix}
  \textbf{S} \textbf{X} = \textbf{R},
\end{equation}
where the matrix $\textbf{S}$ is an $n \times n$ matrix consisting
of 1 along the diagonal and -1 along the subdiagonal,
\begin{equation}
\textbf{S} \equiv \left(
\begin{array}{rrrrr}
1 & 0 & 0 & \cdots & 0 \\
-1 & 1 & 0 & \ddots & \vdots \\
0 & -1 & 1 & \ddots & 0 \\
\vdots & \ddots & \ddots & \ddots & 0 \\
0 & \cdots & 0 & -1 & 1 \\
\end{array}
\right),
\end{equation}
and the matrix $\textbf{R}$ is an $n \times d$ matrix where every column is
a sample from $\mathcal{P}$. Thus $\textbf{X} = \textbf{S}^{-1} \textbf{R}$.

To perform PCA, we need to compute the eigenvalues and eigenvectors of the
covariance matrix $\hat{\textbf{X}}^T \hat{\textbf{X}}$, where
$\hat{\textbf{X}}$ is the matrix $\textbf{X}$ with the mean of every
dimension across all steps subtracted.  $\hat{\textbf{X}}$ can be found by
applying the $n \times n$ centering matrix, $\textbf{C}$:
\begin{equation}
  \hat{\textbf{X}} = \textbf{C} \textbf{X}, \quad \textbf{C} \equiv
  \textbf{I} - \frac{1}{n} \textbf{1} \textbf{1}^T.
\end{equation}
We now note that the analysis is simplified considerably by instead finding
the eigenvalues and eigenvectors of the matrix $\hat{\textbf{X}}
\hat{\textbf{X}}^T$.  The non-zero eigenvalues of $\hat{\textbf{X}}^T
\hat{\textbf{X}}$ are the same as those of $\hat{\textbf{X}}
\hat{\textbf{X}}^T$.  The eigenvectors are similarly related by
$\textbf{v}_k = \textbf{X}^T \textbf{u}_k$, where $\textbf{v}_k$ is a
(non-normalized) eigenvector of $\hat{\textbf{X}}^T \hat{\textbf{X}}$, and
$\textbf{u}_k$ is the corresponding eigenvector of $\hat{\textbf{X}}
\hat{\textbf{X}}^T$.

We therefore would like to find the eigenvalues and eigenvectors of the matrix
\begin{equation}
\label{eq:pca_eigenvalue_start}
  \hat{\textbf{X}} \hat{\textbf{X}}^T = \textbf{C} \textbf{S}^{-1}
  \textbf{R} \textbf{R}^T \textbf{S}^{-T} \textbf{C},
\end{equation}
where we note that $\textbf{C}^T = \textbf{C}$.  Consider the middle term,
$\textbf{R} \textbf{R}^T$.  In the
limit $d \gg n$ we will have $\textbf{R} \textbf{R}^T \to \textbf{I}$ because the off diagonal terms
will be $\mathbb{E}[\xi_i]^2 = 0$, whereas the diagonal terms will be
$\mathbb{E}[\xi^2] = \sum_{i=0}^d \mathbb{V}[\xi_i] = 1$.  (Recall that we
have assumed that the covariance of the noise distribution is normalized; if
the covariance is not normalized, this simply introduces an overall scale
factor given by the trace of the covariance.)  We therefore have the
simplification
\begin{equation}
  \hat{\textbf{X}} \hat{\textbf{X}}^T = \textbf{C} \textbf{S}^{-1}
  \textbf{S}^{-T} \textbf{C}.
\label{eq X XT}
\end{equation}

\subsection{Asymptotic convergence to circulant matrices}

Let us consider the new middle term, $\textbf{S}^{-1} \textbf{S}^{-T} =
(\textbf{S}^T \textbf{S})^{-1}$.  The matrix $\textbf{S}$ is a banded
Toeplitz matrix.  \citet{gray06} has shown that banded Toeplitz matrices
asymptotically approach circulant matrices as the size of the matrix grows.
In particular, \citet{gray06} showed that banded Toeplitz matrices have the
same inverses, eigenvalues, and eigenvectors as their corresponding
circulant matrices in this asymptotic limit (see especially theorem 4.1 and
subsequent material from \citealt{gray06}).  Thus in our case, if we
consider the limit of a large number of steps, $\textbf{S}$ asymptotically
approaches a circulant matrix $\widetilde{\textbf{S}}$ that is equal to
$\textbf{S}$ in every entry except the top right, where there appears a $-1$
instead of a 0.\footnote{We note in passing that $\widetilde{\textbf{S}}$ is
the exact representation of a \emph{closed} random walk.}

With the circulant limiting behavior of $\textbf{S}$ in mind, the problem
simplifies considerably.  We note that $\textbf{C}$ is also a circulant
matrix, the product of two circulant matrices is circulant, the transpose of
a circulant matrix is circulant, and the inverse of a circulant matrix is
circulant.  Thus the matrix $\hat{\textbf{X}} \hat{\textbf{X}}^T$ is
asymptotically circulant as $n \to \infty$.  Finding the eigenvectors is
trivial because the eigenvectors of all circulant matrices are the Fourier
modes.  To find the eigenvalues we must explicitly consider the values of
$\hat{\textbf{X}} \hat{\textbf{X}}^T$.  The matrix $\textbf{S}^T \textbf{S}$
consists of a 2 along the diagonal, -1 along the subdiagonal and
superdiagonal, and 0 elsewhere, with the exception of the bottom right
corner where there appears a 1 instead of a 2.

While this matrix is not a banded Toeplitz, it is asymptotically equivalent
to a banded Toeplitz matrix because it differs from a banded Toeplitz matrix
by a finite amount in a single location.  We now note that multiplication of
the centering matrix does not change either the eigenvectors or the
eigenvalues of this matrix since all vectors with zero mean are eigenvectors
of the centering matrix with eigenvalue 1, and all Fourier modes but the
first have zero mean.  Thus the eigenvalues of $\hat{\textbf{X}}
\hat{\textbf{X}}^T$ can be determined by the inverse of the non-zero
eigenvalues of $\textbf{S}^T \textbf{S}$, which is an asymptotic circulant
matrix.  The \kth{} eigenvalue of a circulant matrix with entries $c_0, c_1,
\ldots$ in the first row is
\begin{equation}
  \label{eq:circulant_eigenvalue}
  \lambda_{\textrm{{\scriptsize circ}}, k} = c_0 + c_{n-1} \omega_k + c_{n-2}
  \omega_k^2 + \ldots + c_1 \omega_k^{n-1},
\end{equation}
where $\omega_k$ is the \kth{} root of unity.  The imaginary parts of the
roots of unity cancel out, leaving the \kth{} eigenvalue of $\textbf{S}^T
\textbf{S}$ to be
\begin{equation}
  \lambda_{\textbf{S}^T \textbf{S}, k} = 2 \left[ 1 - \cos \left( \frac{\pi
  k}{n} \right) \right],
\end{equation}
and the \kth{} eigenvalue of $\hat{\textbf{X}} \hat{\textbf{X}}^T$ to be
\begin{equation}
  \lambda_{\hat{\textbf{X}} \hat{\textbf{X}}^T, k} = \frac{1}{2} \left[ 1 -
  \cos \left( \frac{\pi k}{n} \right) \right]^{-1}.
\end{equation}
The sum of the eigenvalues is given by the trace of $(\textbf{S}^T
\textbf{S})^{-1} = \textbf{S}^{-1} \textbf{S}^{-T}$, and $\textbf{S}^{-1}$
is given by a lower triangular matrix with ones everywhere on and below the
diagonal.  The trace of $(\textbf{S}^T \textbf{S})^{-1}$ is therefore given
by
\begin{equation}
  \label{eq:trace}
  \textrm{Tr} \left( \textbf{S}^{-1} \textbf{S}^{-T} \right) = \frac{1}{2} n
  (n + 1),
\end{equation}
and so the explained variance ratio from the \kth{} PCA component,
$\rho_k$ in the limit $n \to \infty$ is
\begin{equation}
  \rho_k \equiv \frac{\lambda_k}{\textrm{Tr}\left( \textbf{S}^{-1}
    \textbf{S}^{-T} \right)}
  = \frac{\frac{1}{2} \left[ 1 - \cos \left( \frac{\pi
    k}{n} \right) \right]^{-1}}{\frac{1}{2} n (n + 1)}.
\end{equation}
If we let $n \to \infty$ we can consider only the first term in a Taylor
expansion of the cosine term.  Requiring that $\sum_{k=1}^{\infty} \rho_k =
1$, the explained variance ratio is
\begin{equation}
  \label{eq:variance_ratio}
  \rho_k = \frac{6}{\pi^2 k^2}.
\end{equation}
We test Eq.~\ref{eq:variance_ratio} empirically in
Fig.~\ref{fig:variance_ratio} in the supplementary material.

We pause here to marvel that the explained variance ratio of a random walk
in the limit of infinite dimensions is \emph{highly} skewed towards the
first few PCA components.  Roughly 60\% of the variance is explained by the
first component, $\sim$80\% by the first two components, $\sim$95\% by the
first 12 components, and $\sim$99\% by the first 66 components.

\subsection{Projection of the trajectory onto PCA components}

Let us now turn to the trajectory of the random walk when projected onto the
PCA components.  The trajectory projected onto the \kth{} PCA component is
\begin{equation}
  \textbf{X}_{\textrm{{\scriptsize PCA}}, k} = \textbf{X} \hat{\textbf{v}}_k,
\end{equation}
where $\hat{\textbf{v}}$ is the normalized $\textbf{v}_k$.  We ignore the
centering operation from here on because it changes neither the eigenvectors
nor the eigenvalues.  From above, we then have
\begin{equation}
  \XPCA = \frac{1}{\|\textbf{v}_k\|} \textbf{X} \textbf{v}_k =
  \frac{1}{\|\textbf{v}_k\|} \textbf{X} \textbf{X}^T \textbf{u}_k
  = \frac{\lambda_k}{\|\textbf{v}_k\|} \textbf{u}_k.
\end{equation}
By the symmetry of the eigenvalue equations $\textbf{X} \textbf{X}^T
\textbf{u} = \lambda \textbf{u}$ and $\textbf{X}^T \textbf{X} \textbf{v} =
\lambda \textbf{v}$, it can be shown that
\begin{equation}
  \|\textbf{v}_k\| = \|\textbf{X}^T \textbf{u}_k\| = \sqrt{\lambda}.
\end{equation}
Since $\textbf{u}_k$ is simply the \kth{} Fourier mode, we therefore have
\begin{equation}
  \label{eq:lissajous}
  \XPCA = \sqrt{\frac{2 \lambda_k}{n}} \cos \left(\frac{\pi k t}{n} \right).
\end{equation}

\begin{figure*}
  \centering
  \includegraphics[width=14cm]{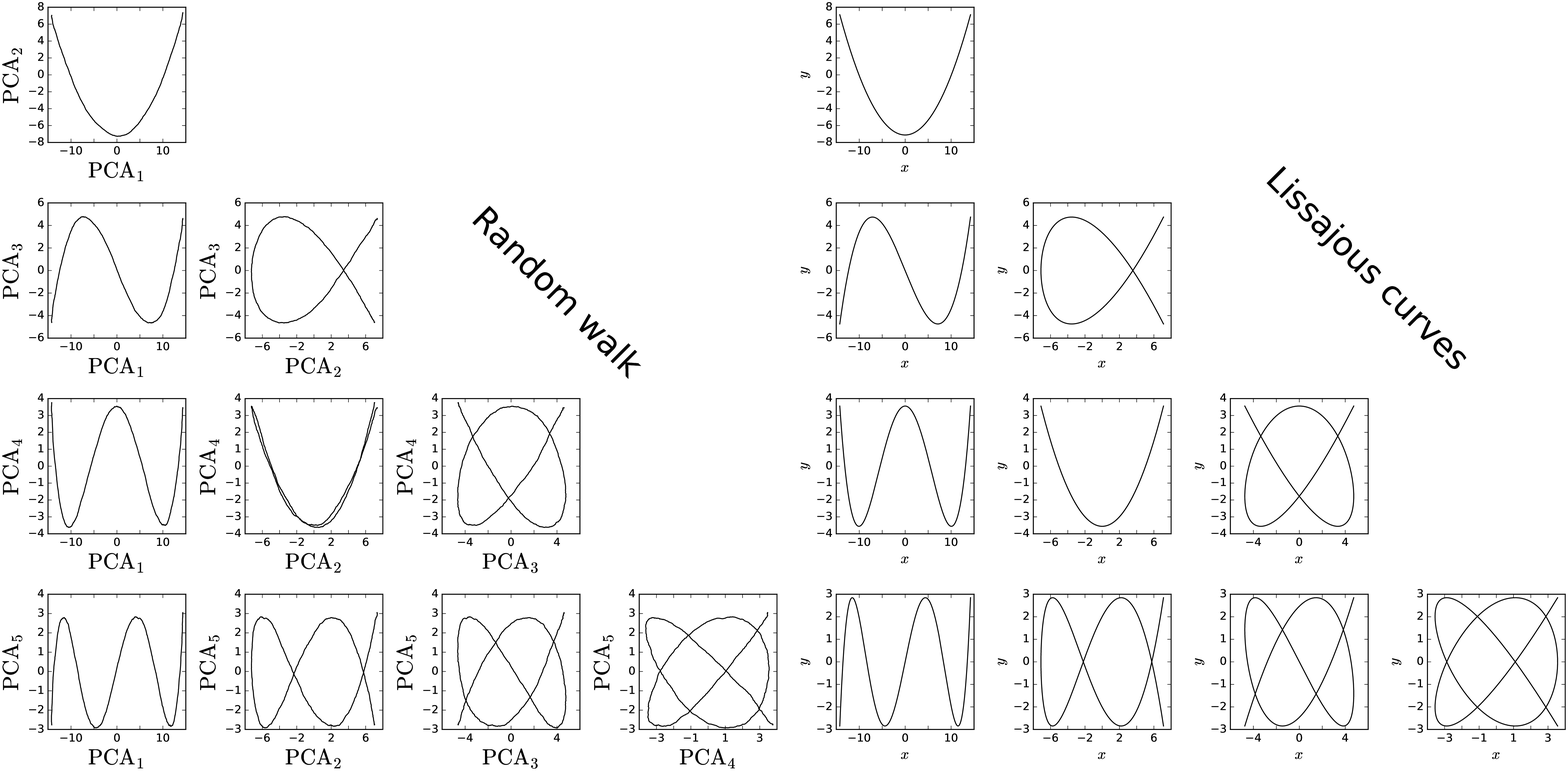}

  \caption{The PCA projections of the trajectories of high dimensional
  random walks are Lissajous curves.  \emph{Left tableau:} Projections of a
  10,000-dimensional random walk onto various PCA components.  \emph{Right
  tableau:} Corresponding Lissajous curves from Eq.~\ref{eq:lissajous}.}

  \label{fig:rw_tableau}
\end{figure*}

This implies that the random walk trajectory projected into the subspace
spanned by two PCA components will be a Lissajous curve.  In
Fig.~\ref{fig:rw_tableau} we plot the trajectories of a high dimensional
random walk projected to various PCA components and compare to the
corresponding Lissajous curves.  We perform 1000 steps of a random walk in
10,000 dimensions and find an excellent correspondence between the empirical
and analytic trajectories.  We additionally show the projection onto the
first few PCA components over time in Fig.~\ref{fig:proj_1d} in the
supplementary material.

While our experiments thus far have used an isotropic Gaussian distribution
for ease of computation, we emphasize that these results are completely
general for \emph{any} probability distribution with zero mean and a finite
covariance matrix with rank much larger than the number of steps.  We
include the PCA projections and eigenvalue distributions of random walks
using non-isotropic multivariate Gaussian distributions in
Figs.~\ref{fig:general_cov_vars} and~\ref{fig:general_cov} in the
supplementary material.

\section{Generalizations}

\subsection{Random walk with momentum}

It is a common practice to train neural networks using stochastic gradient
descent with momentum.  It is therefore interesting to examine the case of a
random walk with momentum.  In this case, the process is governed by the
following set of updates:
\begin{eqnarray}
  \textbf{v}_t & = & \gamma \textbf{v}_{t-1} + \bm{\xi}_t \\
  \textbf{x}_t & = & \textbf{x}_{t-1} + \textbf{v}_t.
\end{eqnarray}
It can be seen that this modifies Eq.~\ref{eq:rw_matrix} to instead read
\begin{equation}
  \textbf{S} \textbf{X} = \textbf{M} \textbf{R}
\end{equation}
where $\textbf{M}$ is a lower triangular Toeplitz matrix with 1 on the
diagonal and $\gamma^k$ on the \kth{} subdiagonal.  The analysis from
Section~\ref{sec:flat_space} is unchanged, except that now instead of
considering the matrix $\textbf{S}^{-1} \textbf{S}^{-T}$ we have the matrix
$\textbf{S}^{-1} \textbf{M} \textbf{M}^T \textbf{S}^{-T}$.  Although
$\textbf{M}$ is not a banded Toeplitz matrix, its terms decay exponentially
to zero for terms very far from the main diagonal. It is therefore
asymptotically circulant as well, and the eigenvectors remain Fourier modes.
To find the eigenvalues consider the product $(\textbf{S}^T \textbf{M}^{-T}
\textbf{M}^{-1} \textbf{S})^{-1}$, noting that $\textbf{M}^{-1}$ is a matrix
with 1s along the main diagonal and $-\gamma$s subdiagonal.  With some
tedious calculation it can be seen that the matrix $\textbf{S}^T
\textbf{M}^{-T} \textbf{M}^{-1} \textbf{S}$ is given by
\begin{equation}
  (\textbf{S} \textbf{M}^{-1} \textbf{M}^{-T} \textbf{S}^T)_{ij} = \left\{
  \begin{array}{ll}
    2 + 2 \gamma + \gamma^2, & i = j \\
    -(1 + \gamma)^2, & i = j \pm 1 \\
    \gamma, & i = j \pm 2 \\
    0, & \textrm{otherwise}
  \end{array}\right.
\end{equation}
with the exception that $\textbf{S}_{nn} = 1$, and $\textbf{S}_{n, n-1} =
\textbf{S}_{n-1, n} = -(1 + \gamma)$.  As before, this matrix is
asymptotically circulant, so the eigenvalues of its inverse are
\begin{equation}
  \label{eq:momentum_eigenvalues}
  \lambda_k = \frac{1}{2} \left[ 1 + \gamma + \gamma^2
      - (1 + \gamma)^2 \cos \left( \frac{\pi k}{n} \right)
      + \gamma \cos \left( \frac{2 \pi k}{n} \right) \right]^{-1}.
\end{equation}
In the limit of $n \to \infty$, the distribution of eigenvalues is identical
to that of a random walk in flat space, however for finite $n$, it has the
effect of shifting the distribution towards the lower PCA components.  We
empirically test Eq.~\ref{eq:momentum_eigenvalues} in
Fig.~\ref{fig:momentum_eigenvalues} in the supplementary material.

\subsection{Discrete Ornstein-Uhlenbeck processes}

A useful generalization of the above analysis of random walks in flat space is
to consider random walks in a quadratic potential, also known as an AR(1)
process or a discrete Ornstein-Uhlenbeck process. For simplicity we will assume
that the potential has its minimum at the origin. Now every step consists of a
stochastic component and a deterministic component which points toward the
origin and is proportional in magnitude to the distance from the origin.  In
this case the update equation can be written
\begin{equation}
  \textbf{x}_t = (1 - \alpha) \textbf{x}_{t - 1} + \bm{\xi}_t,
\end{equation}
where $\alpha$ measures the strength of the potential.  In the limit $\alpha
\to 0$ the potential disappears and we recover a random walk in flat space.
In the limit $\alpha \to 1$ the potential becomes infinitely strong and we
recover independent samples from a multivariate Gaussian distribution.  For
$1 < \alpha < 2$ the steps will oscillate across the origin.  For $\alpha$
outside $[0, 2]$ the updates diverge exponentially.

\subsubsection{Analysis of eigenvectors and eigenvalues}

This analysis proceeds similarly to the analysis in
Section~\ref{sec:flat_space} except that instead of $\textbf{S}$ we now have
the matrix $\SOU$ which has 1s along the diagonal and $-(1 - \alpha)$ along
the subdiagonal.  $\SOU$ remains a banded Toeplitz matrix and so the
arguments from Sec.~\ref{sec:flat_space} that $\hat{\textbf{\textbf{X}}}
\hat{X}^T$ is asymptotically circulant hold.  This implies that the
eigenvectors of $\hat{\textbf{X}} \hat{\textbf{X}}^T$ are still Fourier
modes.  The eigenvalues will differ, however, because we now have that the
components of $\SOU^T \SOU$ are given by
\begin{equation}
\left( \SOU^T \SOU \right)_{ij} =
\left\{\begin{array}{ll}
  1 + (1 - \alpha)^2, & i < n, i = j \\
   -(1 - \alpha), & i = j \pm 1 \\
  1, & i = j = n \\
   0, & \textrm{otherwise.} \\
\end{array}\right.
\end{equation}
From Eq.~\ref{eq:circulant_eigenvalue} we have that the \kth{} eigenvalue of
$\SOU^T \SOU$ is
\begin{equation}
  \label{eq:ou_eigenvalue}
  \lambda_{\textrm{{\scriptsize OU}}, k} = \left[ 1 + (1 - \alpha)^2 -
  2(1 - \alpha) \cos \left( \frac{2 \pi k}{n} \right) \right]^{-1}
  \simeq \left[ \frac{4 \pi^2 k^2 (1 - \alpha)}{n^2} + \alpha^2
  \right]^{-1}.
\end{equation}

\begin{figure}
  \centering
  \includegraphics[width=14cm]{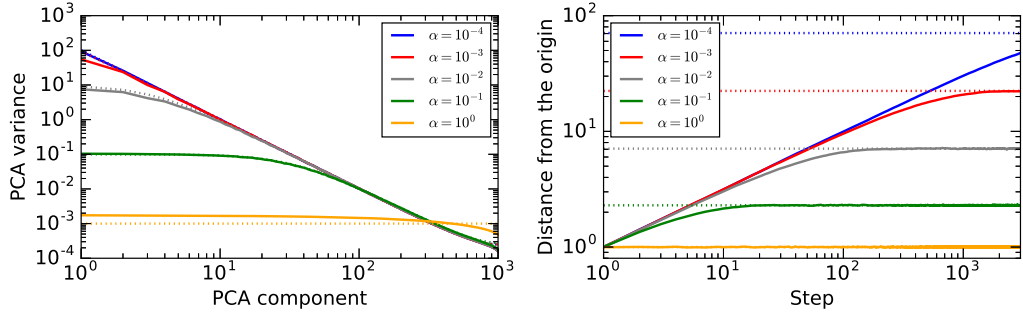}

  \caption{\emph{Left panel:} The variance of the PCA components for several
  choices of $\alpha$.  The empirical distribution is shown in solid and the
  predicted distribution with a dotted line.  The predicted distribution
  generally matches the observed distribution closely, but there is a systematic
  deviation for $\alpha$ near 1.  This is due to the fact that when the mean
  distribution is flat, there will nevertheless be a distribution around this
  mean when these eigenvalues are sampled from real data.  Because PCA sorts
  these eigenvalues, this will always lead to a deviation from the flat
  distribution.  \emph{Right panel:} Distance from the origin for discrete
  Ornstein-Uhlenbeck processes with several choices of $\alpha$ (solid lines)
  with the predicted asymptote from Eq.~\ref{eq:critical_radius} (dotted
  lines).}

  \label{fig:ou_variances_sphere}
\end{figure}

We show in Fig.~\ref{fig:ou_variances_sphere} a comparison between the
eigenvalue distribution predicted from Eq.~\ref{eq:ou_eigenvalue} and the
observed distribution from a 3000 step Ornstein-Uhlenbeck process in 30,000
dimensions for several values of $\alpha$.  There is generally an extremely
tight correspondence between the two.  The exception is in the limit of
$\alpha \to 1$, where there is a catch which we have hitherto neglected.
While it is true that the \emph{mean} eigenvalue of any eigenvector
approaches the same constant, there is nevertheless going to be some
distribution of eigenvalues for any finite walk. Because PCA \emph{sorts}
the eigenvalues, there will be a characteristic deviation from a flat
distribution.

\subsubsection{Critical distance and mixing time}

While we might be tempted to take the limit $n \to \infty$ as we did in the
case of a random walk in flat space, doing so would obscure interesting
dynamics early in the walk.  (A random walk in flat space is self-similar so
we lose no information by taking this limit.  This is no longer the case in
an Ornstein- Uhlenbeck process because the parameter $\alpha$ sets a
characteristic scale in the system.)  In fact there will be two distinct
phases of a high dimensional Ornstein-Uhlenbeck process initialized at the
origin.  In the first phase the process will behave as a random walk in flat
space --- the distance from the origin will increase proportionally to
$\sqrt{n}$ and the variance of the \kth{} PCA component will be proportional
to $k^{-2}$.   However, once the distance from the origin reaches a critical
value, the gradient toward the origin will become large enough to balance
the tendency of the random walk to drift away from the
origin.\footnote{Assuming we start close to the origin.  If we start
sufficiently far from the origin the trajectory will exponentially decay to
this critical value.}  At this point the trajectory will wander indefinitely
around a sphere centered at the origin with radius given by this critical
distance.  Thus, while an Ornstein-Uhlenbeck process is mean-reverting in
low dimensions, in the limit of infinite dimensions the Ornstein-Uhlenbeck
process is no longer mean-reverting --- an infinite dimensional
Ornstein-Uhlenbeck process will never return to its
mean.\footnote{Specifically, since the limiting distribution is a
$d$-dimensional Gaussian, the probability that the process will return to
within $\epsilon$ of the origin is $P(d/2, \epsilon^2/2)$, where $P$ is the
regularized gamma function.  For small $\epsilon$ this decays exponentially
with $d$.}  This critical distance can be calculated by noting that each
dimension is independent of every other and it is well known that the
asymptotic distribution of an AR(1) process with Gaussian noise is Gaussian
with a mean of zero and a standard deviation of $\sqrt{V/(1 - (1 -
\alpha)^2)}$, where $V$ is the variance of the stochastic component of the
process.  In high dimensions the asymptotic distribution as $n \to \infty$
is simply a multidimensional isotropic Gaussian.  Because we are assuming $V
= 1/d$, the overwhelming majority of points sampled from this distribution
will be in a narrow annulus at a distance
\begin{equation}
  \label{eq:critical_radius}
  r_c = \frac{1}{\sqrt{\alpha (2 - \alpha)}}
\end{equation}
from the origin.  Since the distance from the origin during the initial
random walk phase grows as $\sqrt{n}$, the process will start to deviate
from a random walk after $n_c \sim (\alpha (2 - \alpha))^{-1}$ steps.  We
show in the right panel of Fig.~\ref{fig:ou_variances_sphere} the distance
from the origin over time for 3000 steps of Ornstein-Uhlenbeck processes in
30,000 dimensions with several different choices of $\alpha$.  We compare to
the prediction of Eq.~\ref{eq:critical_radius} and find a good match.

\subsubsection{Iterate averages converge slowly}

We finally note that if the location of the minimum is unknown, then iterate
(or Polyak) averaging can be used to provide a better estimate.  But the
number of steps must be much greater than $n_c$ before iterate averaging
will improve the estimate.  Only then will the location on the sphere be
approximately orthogonal to its original location on the sphere and the
variance on the estimate of the minimum will decrease as $1 / \sqrt{n}$.  We
compute the mean of converged Ornstein-Uhlenbeck processes with various
choices of $\alpha$ in Fig.~\ref{fig:polyak} in the supplementary material.

\subsubsection{Random walks in non-isotropic potential are dominated by low
curvature directions}

While our analysis has been focused on the special case of a quadratic
potential with equal curvature in all dimensions, a more realistic quadratic
potential will have a distribution of curvatures and the axes of the
potential may not be aligned with the coordinate basis.  Fortunately these
complications do not change the overall picture much.  For a general
quadratic potential described by a positive semi-definite matrix $A$, we can
decompose $A$ into its eigenvalues and eigenvectors.  We then apply a
coordinate transformation to align the parameter space with the eigenvectors
of $A$.  At this point we have a distribution of curvatures, each one given
by an eigenvalue of $A$.  However, because we are considering the limit of
infinite dimensions, we can assume that there will be a large number of
dimensions that fall in any bin $[\alpha_i, \alpha_i + d\alpha]$. Each of
these bins can be treated as an independent high-dimensional
Ornstein-Uhlenbeck process with curvature $\alpha_i$.  After $n$ steps, PCA
will then be dominated by dimensions for which $\alpha_i$ is small enough
that $n \ll n_{c, i}$.  Thus, even if relatively few dimensions have small
curvature they will come to dominate the PCA projected trajectory after
enough steps.

\section{Comparison to linear models and neural networks}

While random walks and Ornstein-Uhlenbeck processes are analytically
tractable, there are several important differences between these simple
processes and optimization of even linear models.  In particular, the
statistics of the noise will depend on the location in parameter space and
so will change over the course of training.  Furthermore, there may be
finite data or finite trajectory length effects.

\begin{figure}
  \centering
  \includegraphics[width=14cm]{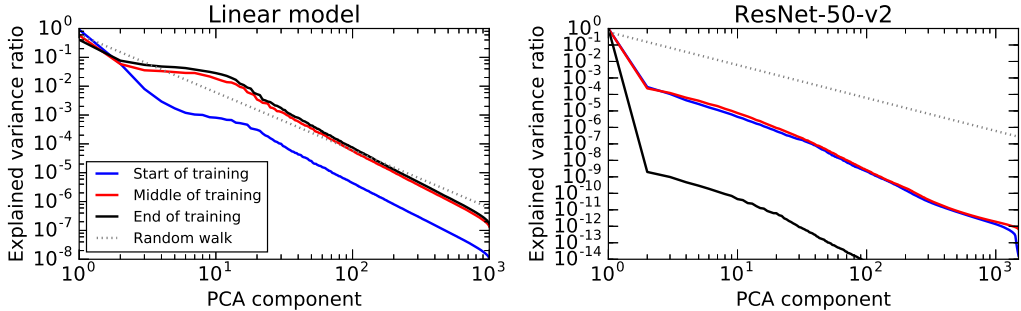}

  \caption{\emph{Left panel:} The distribution of PCA variances at various
  points in training for a linear model trained on CIFAR-10.  At the beginning
  of training the model's trajectory is more directed than a random walk, as
  exhibited by the steep distribution in the lower PCA components.  By the
  middle of training this distribution has flattened (apart from the first PCA
  component) and more closely resembles that of an Ornstein-Uhlenbeck process.
  \emph{Right panel:} The distribution of PCA variances of the parameters of
  ResNet-50-v2 at various points in training.  The distribution of PCA variances
  generally matches that of a random walk with the exception of the first PCA
  component, which dominates the distribution, particularly at the end of
  training.}

  \label{fig:model_pca}
\end{figure}

To get a sense for the effect of these differences we now compare the
distribution of the variances in the PCA components between two models and a
random walk.  For our first model we train a linear model without biases on
CIFAR-10 using a learning rate of $10^{-5}$ for 10,000 steps. For our second
model we train ResNet-50-v2 on Imagenet without batch normalization for
150,000 steps using SGD with momentum and linear learning rate decay.  We
collect the value of all parameters at every step for the first 1500 steps,
the middle 1500 steps, and the last 1500 steps of training, along with
collecting the parameters every 100 steps throughout the entirety of
training.  Further details of both models and the training procedures can be
found in the supplementary material.  While PCA is tractable on a linear
model of CIFAR-10, ResNet-50-v2 has $\sim$25 million parameters and
performing PCA directly on the parameters is infeasible, so we instead
perform a random Gaussian projection into a subspace of 30,000 dimensions.
We show in Fig.~\ref{fig:model_pca} the distribution of the PCA variances at
the beginning, middle, and end of training for both models and compare to
the distribution of variances from an infinite dimensional random walk.  We
show tableaux of the PCA projected trajectories from the middle of training
for the linear model and ResNet-50-v2 in Fig.~\ref{fig:mid_tableaux}.
Tableaux of the other training trajectories in various PCA subspaces are
shown in the supplementary material.

The distribution of eigenvalues of the linear model resembles an OU process,
whereas the distribution of eigenvalues of ResNet-50-v2 resembles a random
walk with a large drift term.  The trajectories appear almost identical to
those of random walks shown in Fig.~\ref{fig:rw_tableau}, with the exception
that there is more variance along the first PCA component than in the random
walk case, particularly at the start and end points.  This manifests itself
in a small outward turn of the edges of the parabola in the PCA2 vs.~PCA1
projection.  This suggests that ResNet-50-v2 generally moves in a consistent
direction over relatively long spans of training, similarly to an
Ornstein-Uhlenbeck process initialized beyond $r_c$.

\begin{figure}
  \centering
  \includegraphics[width=14cm]{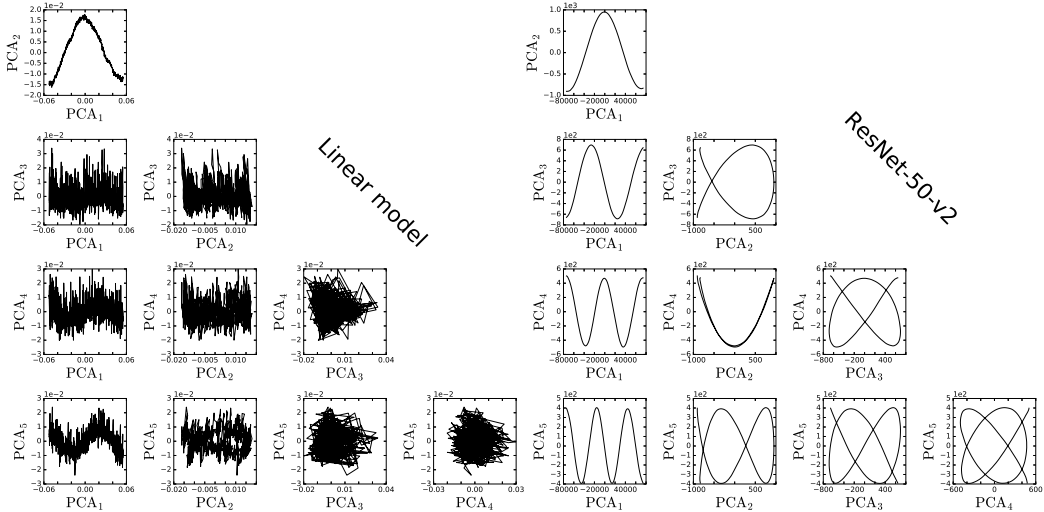}

  \caption{\emph{Left tableau:} PCA projected trajectories from the middle
  of training a linear model on CIFAR-10.  Training has largely converged at
  this point, producing an approximately Gaussian distribution in the higher
  PCA components.  \emph{Right tableau:} PCA projected trajectories from the
  middle of training ResNet-50-v2 on Imagenet.  These trajectories strongly
  resemble those of a random walk.  See
  Figs.~\ref{fig:linear_model_tableaux} and~\ref{fig:resnet_tableaux} in the
  supplementary material for PCA projected trajectories at other phases of
  training.}

  \label{fig:mid_tableaux}
\end{figure}

\section{Random walks with decaying step sizes}

\begin{figure}
  \centering
  \includegraphics[width=8cm]{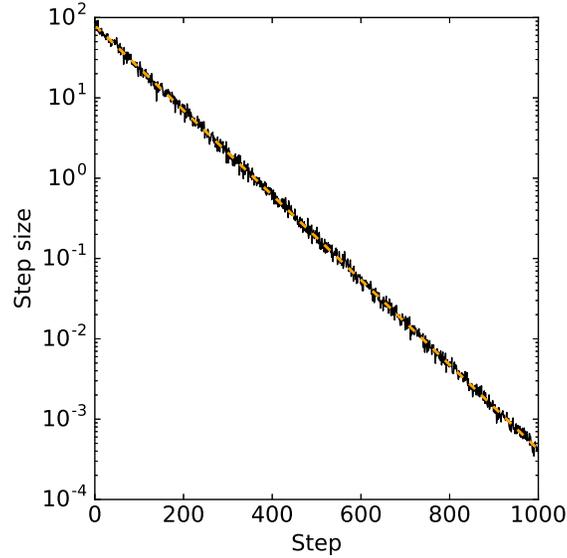}

  \caption{The change in the step size from training a linear model on
  synthetic Gaussian data.  The step size decays exponentially with the best
  fit shown in the orange dashed line.}

  \label{fig:stepsize_decay}
\end{figure}

We finally note that the PCA projected trajectories of the linear model and
ResNet-50-v2 over the entire course of training qualitatively resemble those
of a high dimensional random walk with exponentially decaying step sizes.
To show this we train a linear regression model $y = \textbf{W} \textbf{x}$,
where $\textbf{W}$ is a fixed, unknown vector of dimension 10,000.  We
sample $\textbf{x}$ from a 10,000 dimensional isotropic Gaussian and
calculate the loss
\begin{equation}
  \mathcal{L} = \frac{1}{2}(y - y^{\prime})^2,
\end{equation}
where $y^{\prime}$ is the correct output.  We show in
Fig.~\ref{fig:stepsize_decay} that the step size decays exponentially.  We
fit the decay rate to this data and then perform a random walk in 10,000
dimensions but decay the variance of the stochastic term $\bm{\xi}_i$ by
this rate.  We compare in Fig.~\ref{fig:decay_tableaux} of the supplementary
material the PCA projected trajectories of the linear model trained on
synthetic data to the decayed random walk.  We note that these trajectories
resemble the PCA trajectories over the entire course of training observed in
Figs.~\ref{fig:linear_model_tableaux} and~\ref{fig:resnet_tableaux} for the
linear model trained on CIFAR-10 and ResNet-50-v2 trained on Imagenet.

\section{Conclusions}

We have derived the distribution of the variances of the PCA components of a
random walk both with and without momentum in the limit of infinite
dimensions, and proved that the PCA projections of the trajectory are
Lissajous curves.  We have argued that the PCA projected trajectory of a
random walk in a general quadratic potential will be dominated by the
dimensions with the smallest curvatures where they will appear similar to a
random walk in flat space.  Finally, we find that the PCA projections of the
training trajectory of a layer in ResNet-50-v2 qualitatively resemble those
of a high dimensional random walk despite the many differences between the
optimization of a large NN and a high dimensional random walk.

\subsubsection*{Acknowledgments}

The authors thank Matthew Hoffman, Martin Wattenberg, Jeffrey Pennington,
Roy Frostig, and Niru Maheswaranathan for helpful discussions and comments
on drafts of the manuscript.

\bibliography{refs}

\begin{thebibliography}{14}
\providecommand{\natexlab}[1]{#1}
\providecommand{\url}[1]{\texttt{#1}}
\expandafter\ifx\csname urlstyle\endcsname\relax
  \providecommand{\doi}[1]{doi: #1}\else
  \providecommand{\doi}{doi: \begingroup \urlstyle{rm}\Url}\fi

\bibitem[Ahn et~al.(2012)Ahn, Korattikara, and Welling]{ahn+12}
Ahn, Sungjin, Korattikara, Anoop, and Welling, Max.
\newblock Bayesian posterior sampling via stochastic gradient fisher scoring.
\newblock \emph{arXiv preprint arXiv:1206.6380}, 2012.

\bibitem[Choromanska et~al.(2015)Choromanska, Henaff, Mathieu, Arous, and
  LeCun]{choromanska+15}
Choromanska, Anna, Henaff, Mikael, Mathieu, Michael, Arous, G{\'e}rard~Ben, and
  LeCun, Yann.
\newblock The loss surfaces of multilayer networks.
\newblock In \emph{Artificial Intelligence and Statistics}, pp.\  192--204,
  2015.

\bibitem[Dauphin et~al.(2014)Dauphin, Pascanu, Gulcehre, Cho, Ganguli, and
  Bengio]{dauphin+14}
Dauphin, Yann~N, Pascanu, Razvan, Gulcehre, Caglar, Cho, Kyunghyun, Ganguli,
  Surya, and Bengio, Yoshua.
\newblock Identifying and attacking the saddle point problem in
  high-dimensional non-convex optimization.
\newblock In \emph{Advances in neural information processing systems}, pp.\
  2933--2941, 2014.

\bibitem[Goodfellow et~al.(2014)Goodfellow, Vinyals, and Saxe]{goodfellow+15}
Goodfellow, Ian~J, Vinyals, Oriol, and Saxe, Andrew~M.
\newblock Qualitatively characterizing neural network optimization problems.
\newblock \emph{arXiv preprint arXiv:1412.6544}, 2014.

\bibitem[Gray(2006)]{gray06}
Gray, Robert~M.
\newblock Toeplitz and circulant matrices: A review.
\newblock \emph{Foundations and Trends{\textregistered} in Communications and
  Information Theory}, 2\penalty0 (3):\penalty0 155--239, 2006.

\bibitem[He et~al.(2016)He, Zhang, Ren, and Sun]{he+16}
He, Kaiming, Zhang, Xiangyu, Ren, Shaoqing, and Sun, Jian.
\newblock Deep residual learning for image recognition.
\newblock In \emph{Proceedings of the IEEE conference on computer vision and
  pattern recognition}, pp.\  770--778, 2016.

\bibitem[Jozefowicz et~al.(2016)Jozefowicz, Vinyals, Schuster, Shazeer, and
  Wu]{jozefowicz+16}
Jozefowicz, Rafal, Vinyals, Oriol, Schuster, Mike, Shazeer, Noam, and Wu,
  Yonghui.
\newblock Exploring the limits of language modeling.
\newblock \emph{arXiv preprint arXiv:1602.02410}, 2016.

\bibitem[Li et~al.(2018)Li, Xu, Taylor, Studor, and Goldstein]{li+18}
Li, Hao, Xu, Zheng, Taylor, Gavin, Studor, Christoph, and Goldstein, Tom.
\newblock Visualizing the loss landscape of neural nets.
\newblock In \emph{International Conference on Learning Representations}, 2018.

\bibitem[Lorch(2016)]{lorch16}
Lorch, Eliana.
\newblock Visualizing deep network training trajectories with pca.
\newblock In \emph{ICML Workshop on Visualization for Deep Learning}, 2016.

\bibitem[Mandt et~al.(2016)Mandt, Hoffman, and Blei]{mandt+16}
Mandt, Stephan, Hoffman, Matthew, and Blei, David.
\newblock A variational analysis of stochastic gradient algorithms.
\newblock In \emph{International Conference on Machine Learning}, pp.\
  354--363, 2016.

\bibitem[Moore \& Ahmed(2017)Moore and Ahmed]{moore+ahmed18}
Moore, James and Ahmed, Hasan.
\newblock High dimensional random walks can appear low dimensional: Application
  to influenza h3n2 evolution.
\newblock \emph{arXiv preprint arXiv:1707.09361}, 2017.

\bibitem[Novak et~al.(2018)Novak, Bahri, Abolafia, Pennington, and
  Sohl-Dickstein]{novak2018sensitivity}
Novak, Roman, Bahri, Yasaman, Abolafia, Daniel~A, Pennington, Jeffrey, and
  Sohl-Dickstein, Jascha.
\newblock Sensitivity and generalization in neural networks: an empirical
  study.
\newblock \emph{International Conference on Learning Representations}, 2018.

\bibitem[Smith \& Le(2017)Smith and Le]{smith+le17}
Smith, Samuel~L and Le, Quoc~V.
\newblock A bayesian perspective on generalization and stochastic gradient
  descent.
\newblock In \emph{Proceedings of Second workshop on Bayesian Deep Learning
  (NIPS 2017)}, 2017.

\bibitem[Uhlenbeck \& Ornstein(1930)Uhlenbeck and
  Ornstein]{uhlenbeck+ornstein30}
Uhlenbeck, George~E and Ornstein, Leonard~S.
\newblock On the theory of the brownian motion.
\newblock \emph{Physical review}, 36\penalty0 (5):\penalty0 823, 1930.

\end{thebibliography}
\bibliographystyle{icml2018}

\newpage

\section{Further empirical tests}

\subsection{High dimensional random walks}

\begin{figure}
  \centering
  \includegraphics[width=8cm]{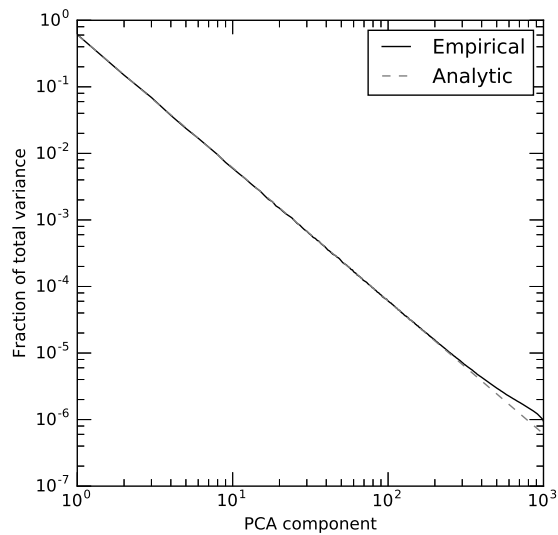}

  \caption{The fraction of the total variance of the different PCA
  components for a high dimensional random walk.  The solid line is
  calculated from performing PCA on a 10,000 dimensional random walk of 1000
  steps.  The dashed line is calculated from the analytic prediction of
  Eq.~\ref{eq:variance_ratio}.  There is excellent agreement up until the
  very largest PCA components where finite size effects start to become
  non-negligible.}

  \label{fig:variance_ratio}
\end{figure}

We test Eq.~\ref{eq:variance_ratio} by computing 1000 steps of a random walk
in 10,000 dimensions and performing PCA on the trajectory.  We show in
Fig.~\ref{fig:variance_ratio} the empirical variance ratio for the various
components compared to the prediction from Eq.~\ref{eq:variance_ratio} and
find excellent agreement.  The empirical variance ratio is slightly higher
than the predicted variance ratio for the highest PCA components due to the
fact that there are a finite number of dimensions in this experiment, so the
contribution from all components greater than the number of steps taken must
be redistributed among the other components, which leads to proportionally
the largest increase in the largest PCA components.

\begin{figure}
  \centering
  \includegraphics[width=8cm]{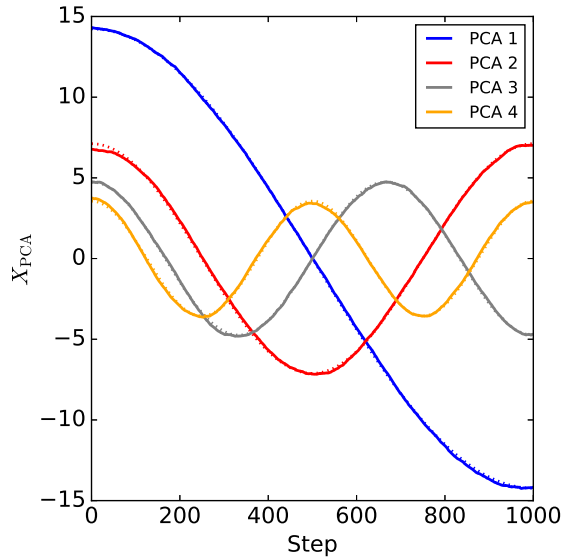}

  \caption{The projection of the trajectory of a high-dimensional random
  walk onto the first five PCA components forms cosines of increasing
  frequency and decreasing amplitude.  The predicted trajectories are shown
  with dotted lines, but the difference between the predicted and observed
  trajectories is generally smaller than the width of the lines.  The random
  walk in this figure consists of 1000 steps in 10,000 dimensions.}

  \label{fig:proj_1d}
\end{figure}

We show in Fig.~\ref{fig:proj_1d} the projection of the trajectory onto the
first few PCA components.  The projection onto the \kth{} PCA component is a
cosine of frequency $k/(2n)$ and amplitude given by
Eq.~\ref{eq:variance_ratio}.

\subsection{Random walk with non-isotropic noise}

\begin{figure}
  \centering
  \includegraphics[width=8cm]{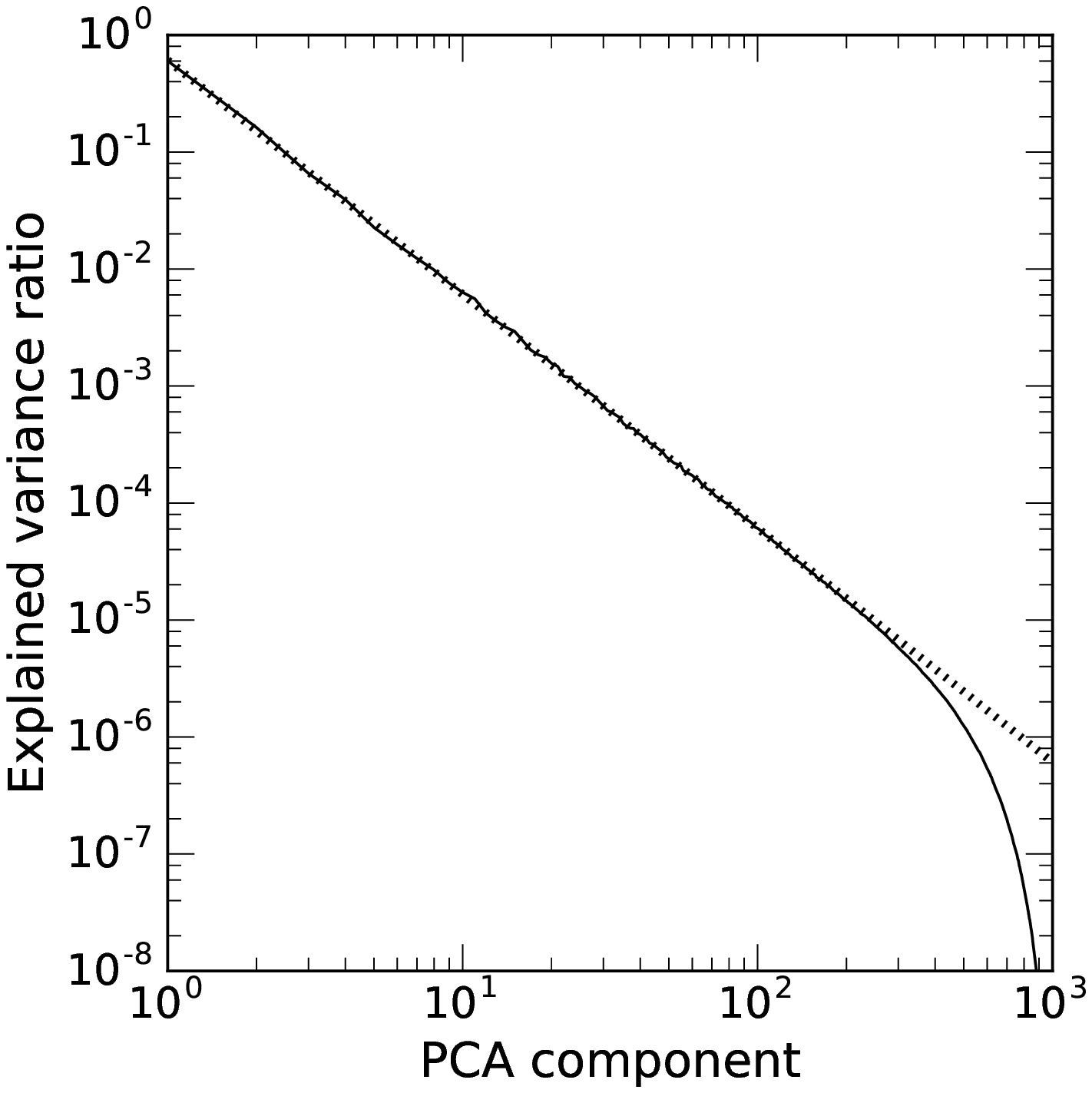}

  \caption{The distribution of explained variance ratios from the PCA of a
  random walk with noise sampled from a multivariate Gaussian with a non-
  isotropic covariance matrix.  Despite the different noise distribution, the
  distribution of explained variance ratios closely matches the prediction from
  Eq.~\ref{eq:variance_ratio}.}

  \label{fig:general_cov_vars}
\end{figure}

\begin{figure}
  \centering
  \includegraphics[width=14cm]{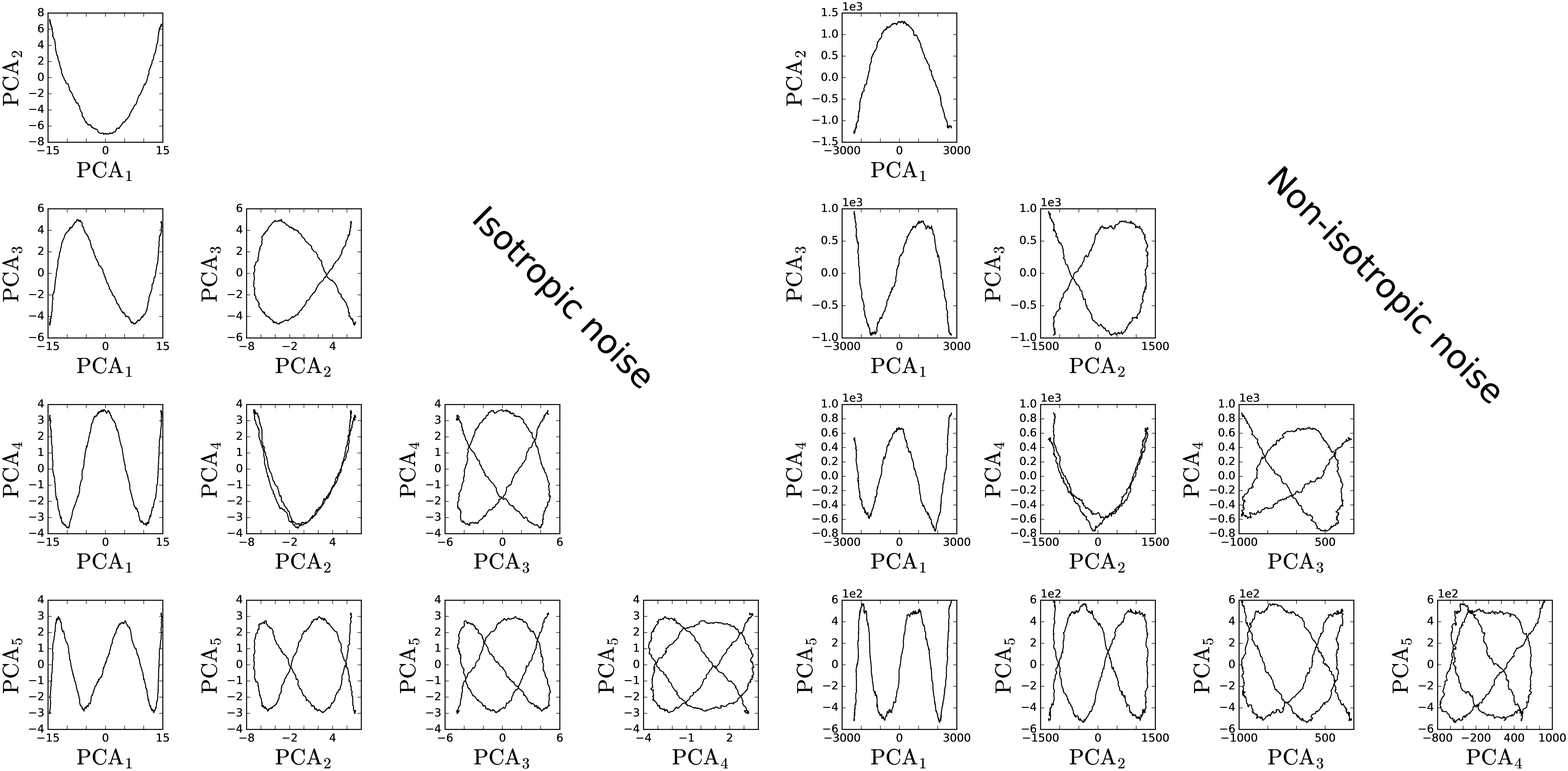}

  \caption{\emph{Left tableau:} The PCA projected trajectory of a random walk
  with noise sampled from an isotropic Gaussian distribution in 1000 dimensions.
  \emph{Right tableau:} The PCA projected trajectory of a random walk with noise
  sampled from a multivariate Gaussian distribution with a random covariance
  matrix in 1000 dimensions.  Although the smaller number of dimensions
  introduces noise into the trajectory, it is clear that the trajectories are
  still Lissajous curves even when the random variates are sampled from a more
  complicated distribution.}

  \label{fig:general_cov}
\end{figure}

To demonstrate that our results hold for non-isotropic noise distributions
we perform a random walk where the noise is sampled from a multivariate
Gaussian distribution with a random covariance matrix, $\bm{\Sigma}$.
Because sampling from a multivariate Gaussian with an arbitrary covariance
matrix is difficult in high dimensions, we restrict the random walk to 1000
dimensions, keeping the number of steps 1000 as before.  To construct the
covariance matrix, we sample a $1000 \times 1000$ dimensional random matrix,
$\textbf{R}$, where each element is a sample from a normal distribution and
then set $\bm{\Sigma} = \textbf{R} \textbf{R}^T$.  Although $\bm{\Sigma}$
will be approximately equal to the identity matrix, the distribution of
eigenvalues will follow a fairly wide Marchenko-Pastur distribution because
$\textbf{R}$ is square.  We show the distribution of explained variance
ratios with the prediction from Eq.~\ref{eq:variance_ratio} in
Fig.~\ref{fig:general_cov_vars}.  There is a tight correspondence between
the two up until the largest PCA components where finite dimension effects
start to dominate.  We also show in Fig.~\ref{fig:general_cov} PCA projected
trajectories of this random walk along with a random walk where the random
variates are sampled from a 1000-dimensional isotropic distribution for
comparison to provide a sense for the amount of noise introduced by the
relatively small number of dimensions.  Although the small dimensionality
introduces noise into the PCA projected trajectories, it is clear that the
general shapes match the predicted Lissajous curves.

\subsection{Random walk with momentum}

\begin{figure}
  \centering
  \includegraphics[width=8cm]{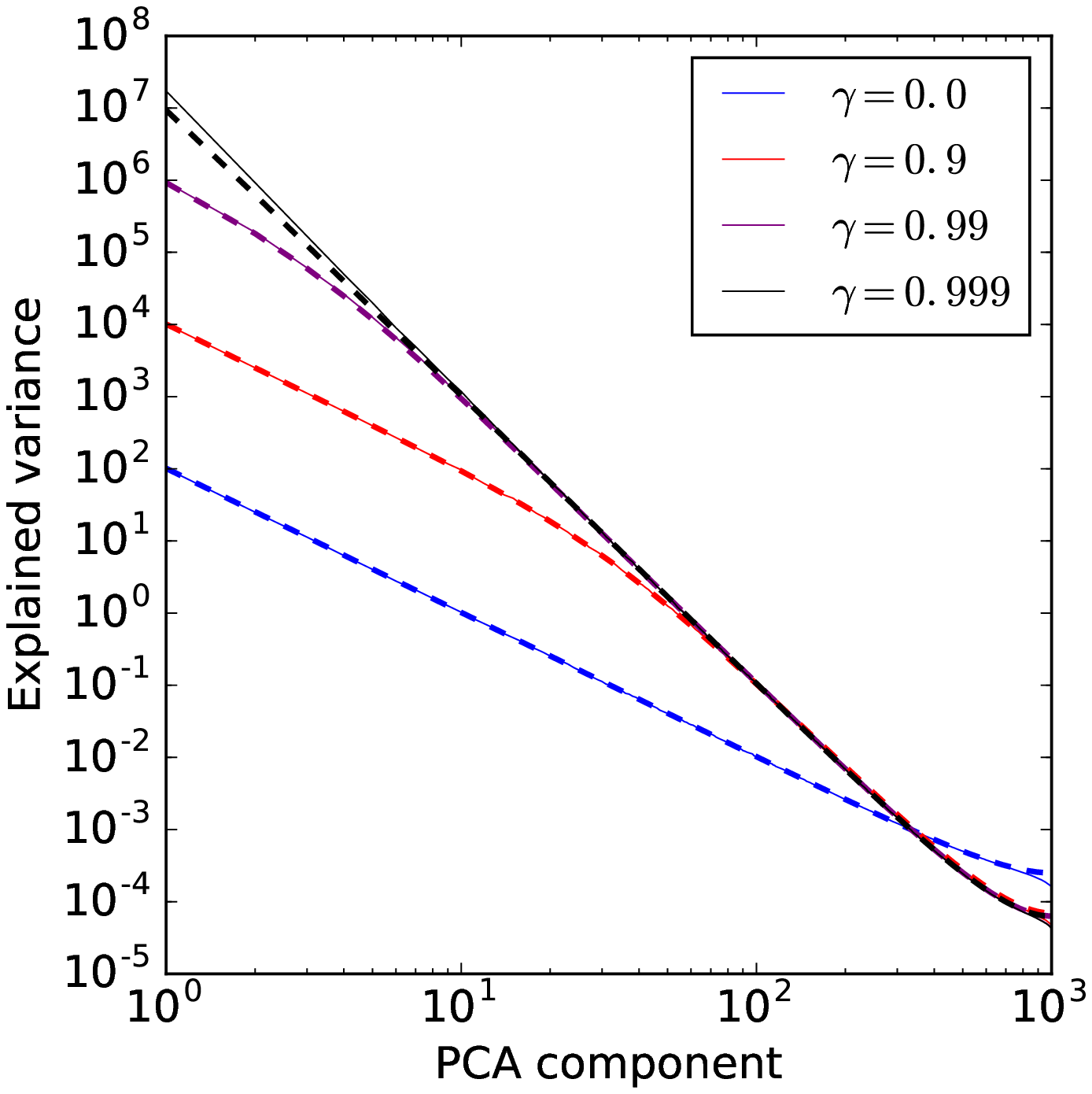}

  \caption{The distribution of explained PCA variances for random
  walks with momentum where we vary the strength of the momentum parameter,
  $\gamma$.  The distribution observed from a 1000 step random walk in
  10,000 dimensions is shown in the solid lines.  The prediction from
  Eq.~\ref{eq:momentum_eigenvalues} is shown in the dashed line.}

  \label{fig:momentum_eigenvalues}
\end{figure}

We test Eq.~\ref{eq:momentum_eigenvalues} by computing 1000 steps of a
random walk in 10,000 dimensions with various choices of the momentum
parameter, $\gamma$.  We show in Fig.~\ref{fig:momentum_eigenvalues} the
observed distribution of PCA variances (not the explained variance ratio)
along with the prediction from Eq.~\ref{eq:momentum_eigenvalues}.  There is
an extremely tight correspondence between the two, except for the lowest PCA
components for $\gamma = 0.999$.  This is expected because the effective
step size is set by $n / (1 - \gamma)$, and because $n = 1000$, the walk
does not have sufficient time to settle into its stationary distribution
of eigenvalues when $\gamma = 0.999$.

\subsection{Iterate averaging of an Ornstein-Uhlenbeck process}

\begin{figure}
  \centering
  \includegraphics[width=8cm]{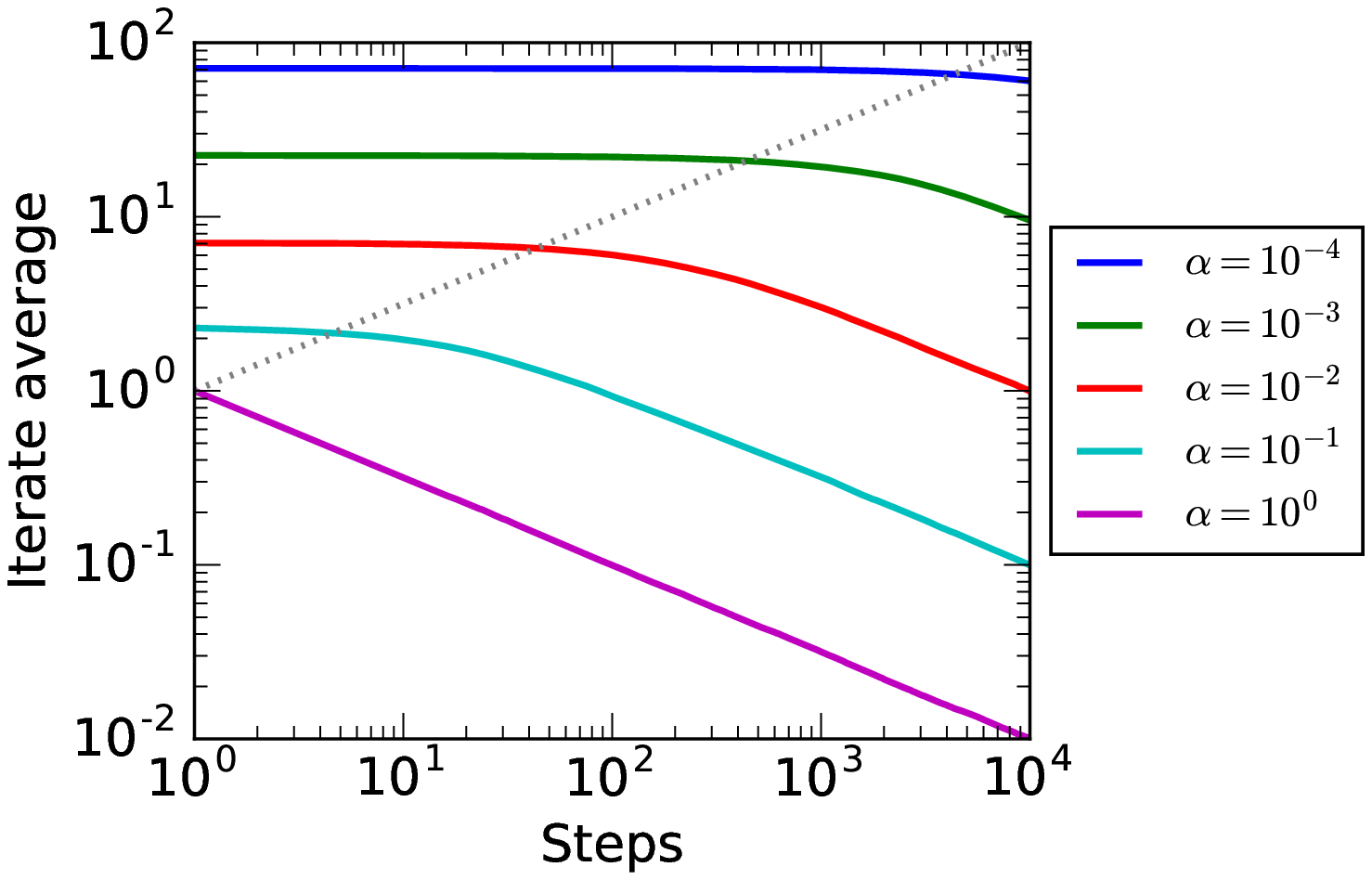}

  \caption{The mean of all steps of converged Ornstein-Uhlenbeck processes of
  various lengths.  The mean remains approximately constant until the total
  angle from the initial position on the sphere grows to $\sim$$\pi/2$, which
  requires $\sim$$\alpha^{-1}$ steps (dashed line).}

  \label{fig:polyak}
\end{figure}

We show in Fig.~\ref{fig:polyak} the mean of all steps of Ornstein-Uhlenbeck
processes which have converged to a random walk on a sphere of radius $r_c$.
We show in the dashed line the predicted value of $n_c$, the number of steps
required to reach $r_c$ (i.e., the crossing time of the sphere).  The
position on the sphere will close to its original location for $n \ll n_c$
so iterate averaging will not improve the estimate of the minimum.  Only
when $n \gg n_c$ will iterate averaging improve the estimate of the minimum
since the correlation between new points wit the original location will be
negligible.

\section{Details of models and training}

\subsection{Linear regression on CIFAR-10}

We train linear regression on CIFAR-10 for 10,000 steps using SGD and a
batch size of 128 and a learning rate of $10^{-5}$.  The model achieves a
validation accuracy of 29.1\%.

\subsection{ResNet-50-v2 on Imagenet}

We train ResNet-50-v2 on Imagenet for 150,000 steps using SGD with momentum
and a batch size of 1024.  We do not use batch normalization since this
could confound our analysis of the training trajectory.  We instead add bias
terms to every convolutional layer.  We decay the learning rate linearly
with an initial learning rate of 0.0345769 to a final learning rate a factor
of 10 lower by 141,553 steps, at which point we keep the learning rate
constant.  We set the momentum to 0.9842.  The network achieves a validation
accuracy of 71.46\%.

\section{Gallery of PCA projected trajectories}

\begin{figure}
  \centering
  \includegraphics[width=14cm]{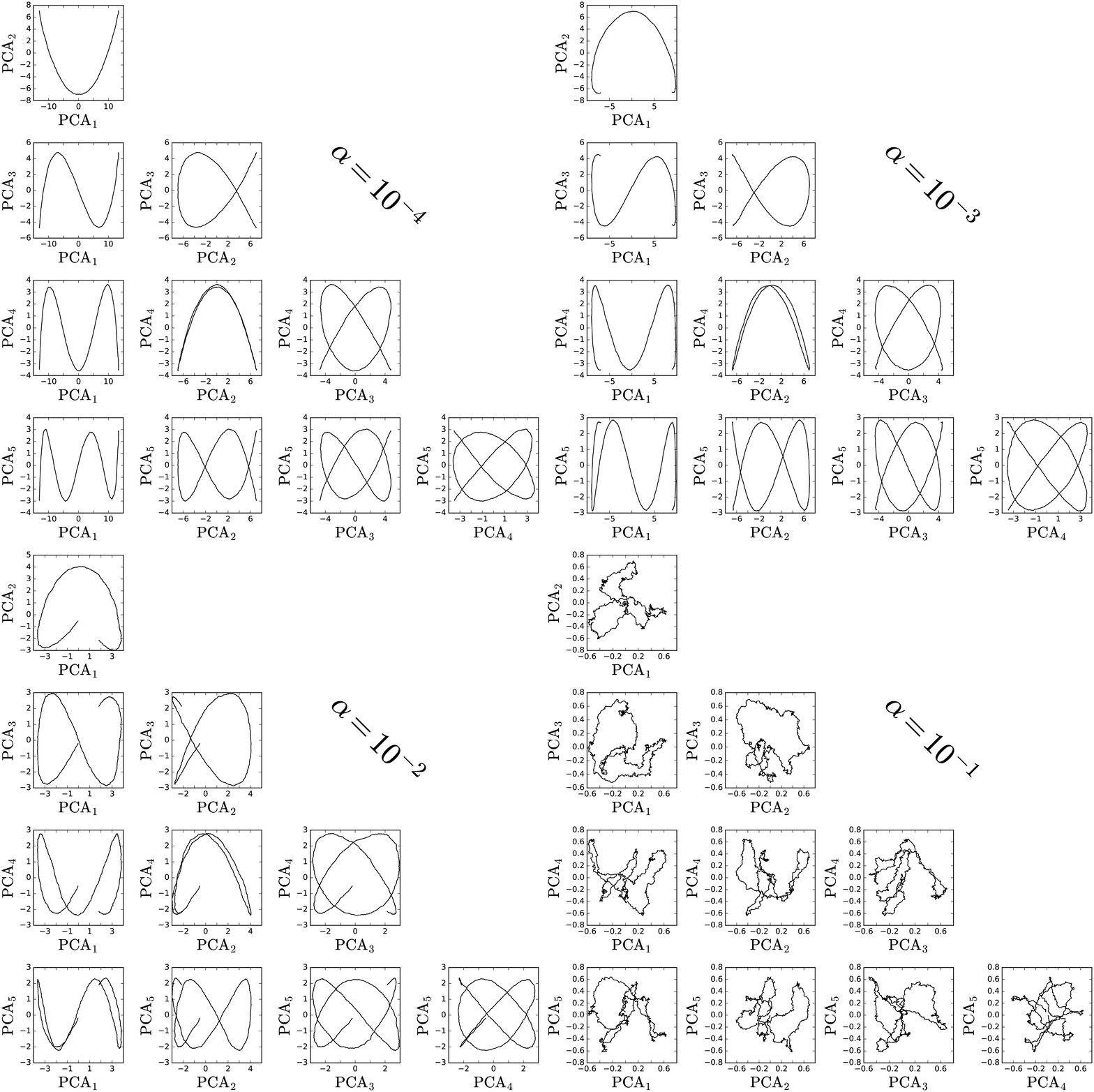}

  \caption{Tableaux of the PCA projections of the trajectories of
  high-dimensional Ornstein-Uhlenbeck processes with various values of
  $\alpha$.  All processes were sampled for 1000 steps in 10,000 dimensions.
  \emph{Upper left tableau:} $\alpha = 10^{-4}$.  \emph{Upper right
  tableau:} $\alpha = 10^{-3}$.  \emph{Lower left tableau:} $\alpha =
  10^{-2}$.  \emph{Lower right tableau:} $\alpha = 10^{-1}$.}

  \label{fig:ou_tableaux}
\end{figure}

We present here tableaux of the PCA projections of various trajectories.  We
show in Fig.~\ref{fig:ou_tableaux} four tableaux of the PCA projections of
the trajectories of high-dimensional Ornstein-Uhlenbeck processes with
different values of $\alpha$.  For $\alpha = 10^{-4}$ the trajectories are
almost identical to a high-dimensional random walk, as they should be since
the process was sampled for only 1000 steps.  Once we have $\alpha^{-1} =
1000$ the trajectories start to visibly deviate from those of a
high-dimensional random walk.  For larger $\alpha$ the deviations continue
to grow until they become unrecognizable at $\alpha = 0.1$ because 1000
steps corresponds to many crossing times on the high dimensional sphere
on which the process takes place.

\begin{figure}
  \centering
  \includegraphics[width=14cm]{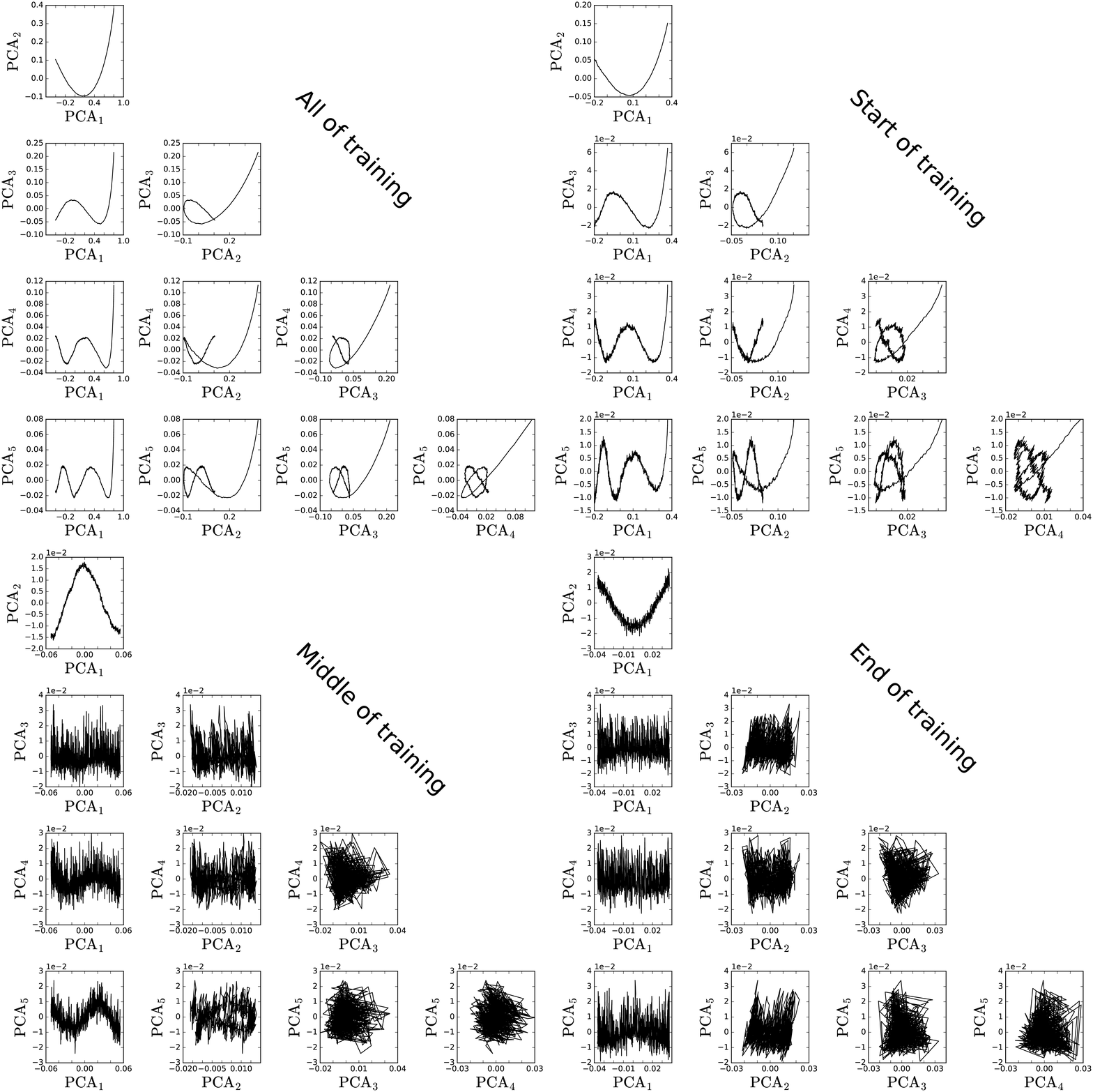}

  \caption{Tableaux of the trajectories of a linear model trained on
  CIFAR-10 in different PCA subspaces.  \emph{Upper left tableau:} PCA
  applied to every tenth step over all of training.  \emph{Upper right
  tableau:} PCA applied to the first 1000 steps of training.  \emph{Lower
  left tableau:} PCA applied to the middle 1000 steps of training.
  \emph{Lower right tableau:} PCA applied to the last 1000 steps of
  training.}

  \label{fig:linear_model_tableaux}
\end{figure}

In Fig.~\ref{fig:linear_model_tableaux} we present tableaux of the PCA
projections of the linear model trained on CIFAR-10.  The trajectory of the
entire training process somewhat resembles a high-dimensional random
walk, though because the model makes larger updates at earlier steps than at
later ones there are long tails on the PCA projected trajectories.  The
model's trajectory most closely resembles a high-dimensional random walk
early in training, but towards the end the higher components become
dominated by noise, implying that these components more closely resemble a
converged Ornstein-Uhlenbeck process.  This corresponds with the flattening
of the distribution of eigenvalues in Fig.~\ref{fig:model_pca}.

\begin{figure}
  \centering
  \includegraphics[width=14cm]{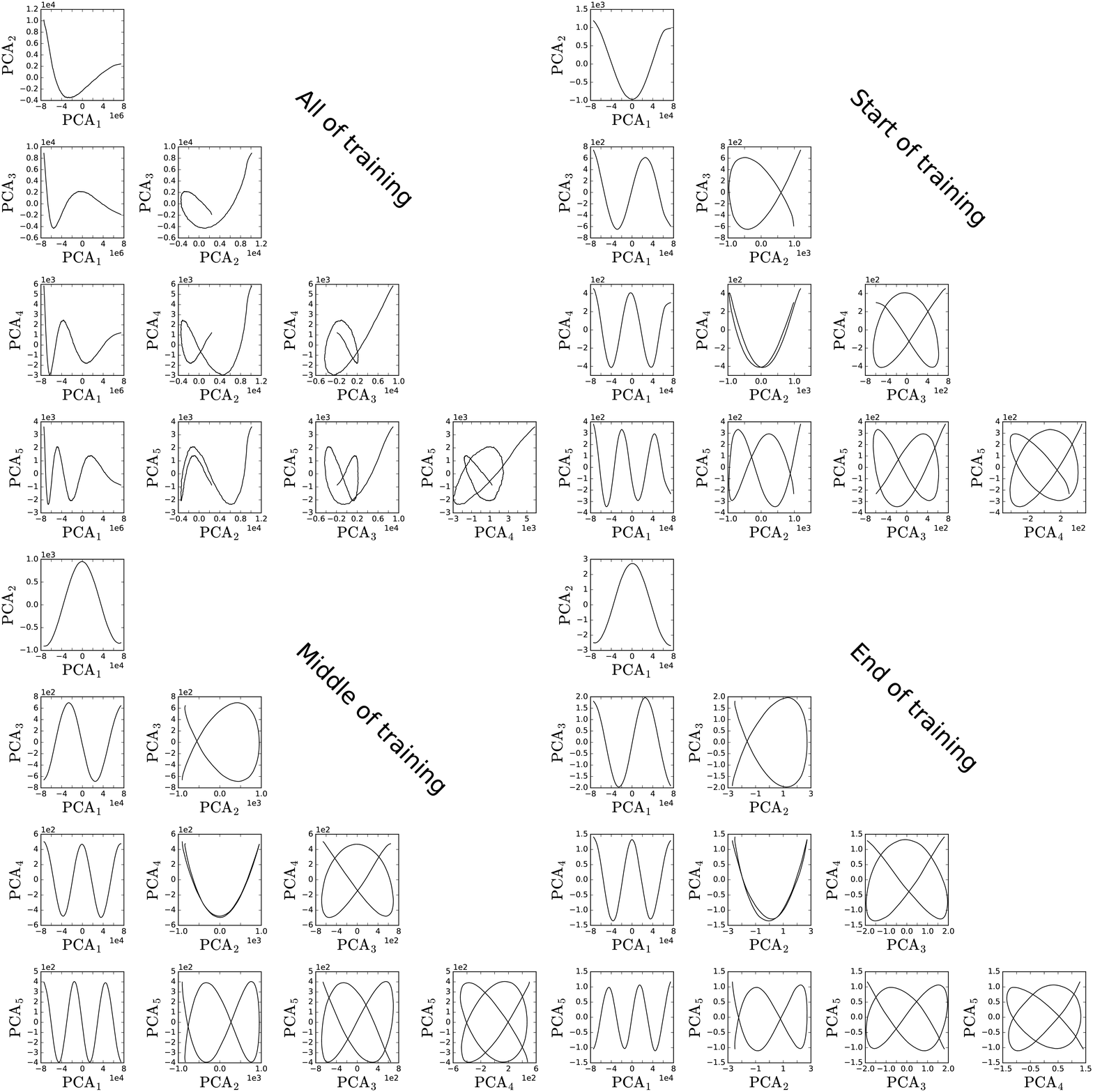}

  \caption{Tableaux of the parameter trajectories of ResNet-50-v2 trained on
  Imagenet in different PCA subspaces.  The parameters were first projected
  into a random Gaussian subspace with 30,000 dimensions before PCA was
  applied.  \emph{Upper left tableau:} PCA applied to every hundredth step
  over all of training.  \emph{Upper right tableau:} PCA applied to the
  first 1500 steps of training.  \emph{Lower left tableau:} PCA applied to
  the middle 1500 steps of training.  \emph{Lower right tableau:} PCA
  applied to the last 1500 steps of training.}

  \label{fig:resnet_tableaux}
\end{figure}

In Fig.~\ref{fig:resnet_tableaux} we present tableaux of the PCA projections
of ResNet-50-v2 trained on Imagenet.  Perhaps remarkably, these trajectories
resemble a high-dimensional random walk much more closely than the linear
model.  However, as in the case of the linear model, the resemblance
deteriorates later in training.

\begin{figure}
  \centering
  \includegraphics[width=14cm]{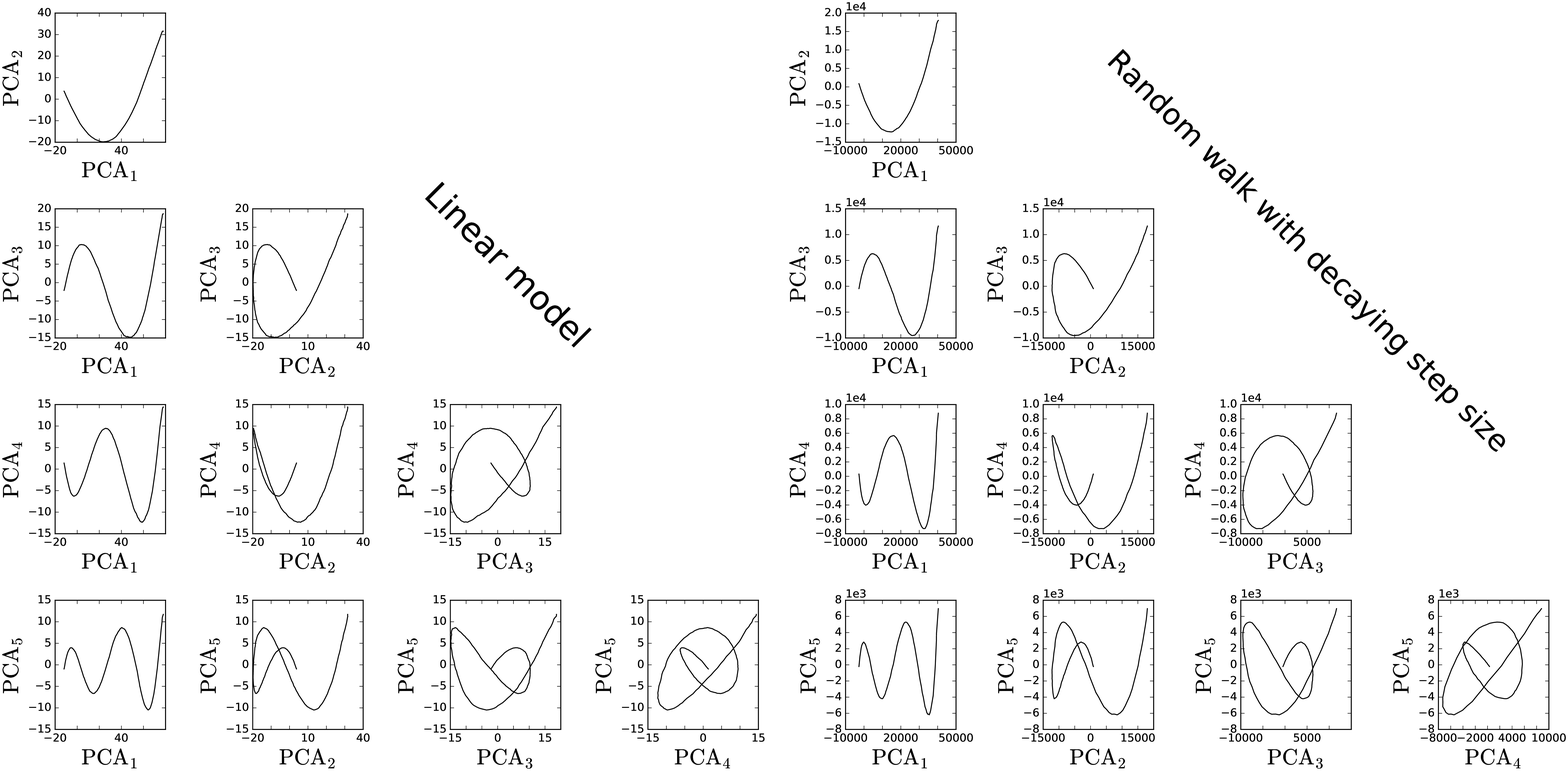}

  \caption{\emph{Left tableau:} PCA projected trajectories of a linear
  regression model trained on synthetic Gaussian data.  \emph{Right
  tableau:} PCA projected trajectories of a 10,000 dimensional random walk
  where the variance of the stochastic component is decayed using the best
  fit found from the linear regression model trained on synthetic data.  The
  trajectories in the two tableaux appear very similar.}

  \label{fig:decay_tableaux}
\end{figure}

\end{document}